\documentclass{article}
\usepackage[utf8]{inputenc}
\usepackage{amssymb}
\usepackage[english]{babel}
\usepackage{amsmath}
\usepackage{graphicx}
\usepackage{bm}
\usepackage{wrapfig}
\usepackage{sidecap}
\usepackage[format=plain, labelfont=it, textfont=it]{caption}
\usepackage[affil-it]{authblk}
\usepackage{cite}
\usepackage{soul, xcolor}

\begin{document}
\setstcolor{red}

\title{Uncertainty Quantification by Ensemble Learning for Computational Optical Form Measurements}
\author{Lara Hoffmann%
  \thanks{Electronic address: \texttt{lara.hoffmann@ptb.de}; Corresponding author}$\ $, Ines Fortmeier and Clemens Elster}
\affil{Physikalisch-Technische Bundesanstalt, Braunschweig and Berlin, Germany}

\date{Dated: \today}
\maketitle
\begin{abstract}
Uncertainty quantification by ensemble learning is explored in terms of an application from computational optical form measurements. The application requires to solve a large-scale, nonlinear inverse problem. Ensemble learning is used to extend a recently developed deep learning approach for this application in order to provide an uncertainty quantification of its predicted solution to the inverse problem. By systematically inserting out-of-distribution errors as well as noisy data the reliability of the developed uncertainty quantification is explored. Results are encouraging and the proposed application exemplifies the ability of ensemble methods to make trustworthy predictions on high dimensional data in a real-world application.
\end{abstract}

\section{Introduction}
Artificial intelligence has established a major impact on science and applications. In particular, deep neural networks \cite{Qin2020} show a great potential of understanding complex scientific relationships through their deep and nonlinear structure. They have been successfully applied to various tasks including natural language processing \cite{Young2018}, computational imaging \cite{Barbastathis2019} or data mining \cite{Wang2020}. 

However, their black-box character and the resulting lack of trustworthiness is probably the most crucial pitfall of deep learning approaches. Many examples exist that demonstrate unreasonable behavior of trained networks. 
For instance, the technique of layer-wise relevance propagation has revealed that an apparently well-trained image classification network actually had adapted a ``Clever Hans" decision strategy \cite{Lapuschkin2019}; the network had learned to classify horse images correctly through focusing at the bottom left corner of the image - there was an unnoticed tag remaining on horse images in the training data set. Adversarial attacks can mislead trained networks to make unreasonable predictions by only slightly perturbing input data \cite{Akhtar2018}. In $2018$ an autonomous driving car crashed into a pedestrian in Arizona because the self-driving system did not classify her correctly \cite{NTSB2019}. 

These examples demonstrate the importance of understanding the behavior of deep neural networks in order to ensure their trustworthiness. Much effort has been spent to develop corresponding approaches. \cite{Adadi2018}, \cite{Lapuschkin2019} and \cite{Selvaraju2017} propose and analyze different methods to explain the behavior of a network, making its predictions more transparent and easier to interpret. In \cite{Martin2020}, the Fisher information is used to detect unusual input to the network. Intense testing is another way of analyzing the behavior of a trained network on critical data and explore its generalization capacity, cf. \cite{Sun2018}, \cite{Tian2018}, \cite{Xu2020}. 

Uncertainty quantification also is an important pillar to improve the trustworthiness of predictions made by a trained network \cite{Kline1985}. Various approaches exist and there are different kinds of uncertainties to consider. Uncertainties are often classified as epistemic and aleatoric (Gal 2017, Hüllermeier \& Waegemann 2019). Sources of uncertainty include imperfect training, unexpected shifts in the data, systematic errors or out-of-distribution data, to mention just a few (cf. \cite{Ashukha2020}, \cite{Kendall2017}, \cite{Ovadia2019}, \cite{Scalia2020}). The most common approaches to uncertainty quantification include Bayesian neural networks \cite{Kononenko1989}, \cite{Yao2019}, dropout based methods \cite{Gal2017}, \cite{Kingma2015} and ensemble techniques \cite{Dietterich2000}, \cite{Lakshminarayanan2017}, \cite{Lee2015}. We focus on the latter, because ensemble learning is straightforward to implement, scales well to higher dimensional data and performs best in recent uncertainty studies \cite{Caldeira2020}, \cite{Gustafsson2020} \cite{Ovadia2019}, \cite{Scalia2020}.

The goal of the paper is to explore the potential of ensemble techniques for uncertainty quantification in deep learning in terms of a large-scale inverse problem from computational optical form measurements. The considered application is based on the tilted-wave interferometer (TWI) which is an accurate, interferometric measurement system for the form measurement of optical aspheres and freeform surfaces  \cite{Baer2014}, \cite{Fortmeier2014}. Applied conventional methods solve the high dimensional, nonlinear, inverse reconstruction problem iteratively through local linearizations. However, the evaluation procedure takes several minutes.

The novelty of the paper is twofold. First, we extend a previous deep learning approach for this application \cite{Hoffmann2020} to incorporate an uncertainty quantification of its predictions. This is achieved through ensemble learning. In contrast to \cite{Hoffmann2020}, the networks are trained on a calibrated data set which will be discussed in more detail later. Second, we systematically insert an increasing out-of-distribution calibration error into the system and analyze its effect on the reliability of the developed uncertainty quantification. Furthermore, the influence of noise is investigated. So far, high-dimensional uncertainty quantification for scalable deep learning techniques is hardly treated in literature\cite{Gustafsson2020} which makes the results for our chosen application interesting also for other machine learning applications.

The paper is structured as follows. Section \ref{section application} introduces the chosen application from computational optical form measurements, followed by a detailed explanation of the data generation in Section \ref{section data generation}. The employed deep neural network, ensemble learning and the corresponding uncertainty quantification are introduced in Section \ref{section method}. Results are then presented in Section \ref{section results} with particular focus on the impact of systematic calibration errors on the uncertainty quantification. In Section \ref{section conclusion}, finally, the potential benefit of deep learning for computational optical form measurements is discussed and possible future research motivated. 

\section{Application}\label{section application}
The application of computational optical form measurements considered here focuses on measuring optical aspheres and freeform surfaces. The application is based on the tilted-wave interferometer (TWI) \cite{Baer2014} which will be introduced in the following.

\begin{figure}[t]
  \centering
  \begin{minipage}[t]{0.36\textwidth}
    \centering
    \includegraphics[scale = 0.6]{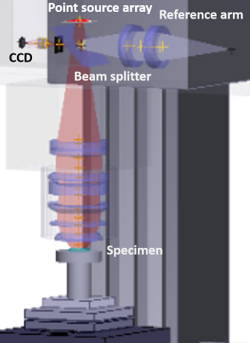}
    \caption{The experimental set-up of the tilted-wave interferometer.}\label{twi}
  \end{minipage}
  \hfill
  \begin{minipage}[t]{0.6\textwidth}
    \centering
    \includegraphics[scale = 0.7]{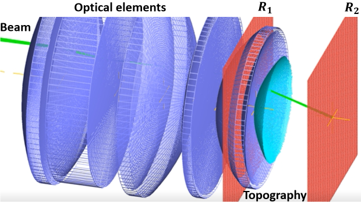}
    \caption{Two reference planes, $R_1$ and $R_2$, are added next to the topography in the computer modeled optical system for calibration.}\label{reference planes}
  \end{minipage}
\end{figure}

The realization of the TWI considered here is the one employed at PTB \cite{Fortmeier2014}, \cite{Fortmeier2017}. Its experimental set-up is shown in Figure \ref{twi}. The coherent light of a laser source (not shown in Fig. 1) is split into a reference and a measurement arm. In the measurement arm, the collimated light passes a 2D micro lens array. Each of these micro lenses acts like a point source thus generating differently tilted wave fronts. After passing through the objective the wavefronts are  reflected at the surface under test and interfere at the beam splitter with the light coming from the reference arm. The resulting intensity images are captured on the charge-coupled device (CCD) and are unwrapped to optical path length differences by using the Goldstein unwrapping algorithm \cite{Goldstein1988}. A beam stop in the Fourier plane of the imaging optics prevents subsampling effects. Therefore, depending on the local slope of the specimen, a different light source generates resolvable sub-interferograms (patches) at the CCD. Information overlap at the CCD is prevented by using four disjoint masks on the point source array, which eventually results in four images of optical path length differences for one specimen topography.

The goal is to measure the deviation of the specimen to its known design topography given the optical path length differences computed from the observed CCD intensities. The toolbox SimOptDevice \cite{Schachtschneider2019} is used to model these optical path length differences in dependence on the topography of the specimen under test. The nonlinear inverse problem consists of finding the topography of the specimen such that the modeled
optical path length differences best fit the observed ones. 

The computer model of the optical system used to solve the inverse problem is not perfect and usually it is ``calibrated''  (i.e. adjusted) using observed data (i.e. optical path length differences) for some test specimens with a high accuracy known topography. The calibration is realized by adding two virtual planes, called \textit{reference planes}, to the computer model of the optical system as shown in Figure \ref{reference planes}. The light beam passes the first plane $R_1$ before attaining the topography and passes the second plane $R_2$ after having attained the topography, respectively. Each virtual reference plane modifies the light beams phenomenologically which is parameterized by Zernike polynomials. Zernike polynomials are orthogonal on the unit disc and commonly used in optics to represent wavefronts \cite{Wang1980}. 
The coefficients of the Zernike polynomials are determined such that the computer model best fits the optical path length differences deduced from the CCD intensities measured by the optical system for a chosen test specimen with a high accuracy known topography \cite{Baer2014_2}. Further details are given in Appendix \ref{appendix calibration}.

\section{Data generation}\label{section data generation}
The goal of the TWI is to measure the deviation of any given specimen to its known design.
We use an asphere as design topography. It is characterized in the Appendix \ref{appendix asphere}. A data set containing various topographies is generated through randomly drawn sets of Zernike coefficients \cite{Wang1980}. The Zernike polynomials parameterize the difference topographies $\Delta T$, i.e. the deviation of the specimen to the known design. Then, the optical path length differences through the optical system are computed for the design topography and each generated specimen, respectively. This is realized with the simulation toolbox SimOptDevice \cite{Schachtschneider2019}. Hence, for each specimen 
a difference of optical path length differences $\Delta L$ is obtained.

\begin{figure}[t]
    \centering
    \includegraphics[scale = 0.33]{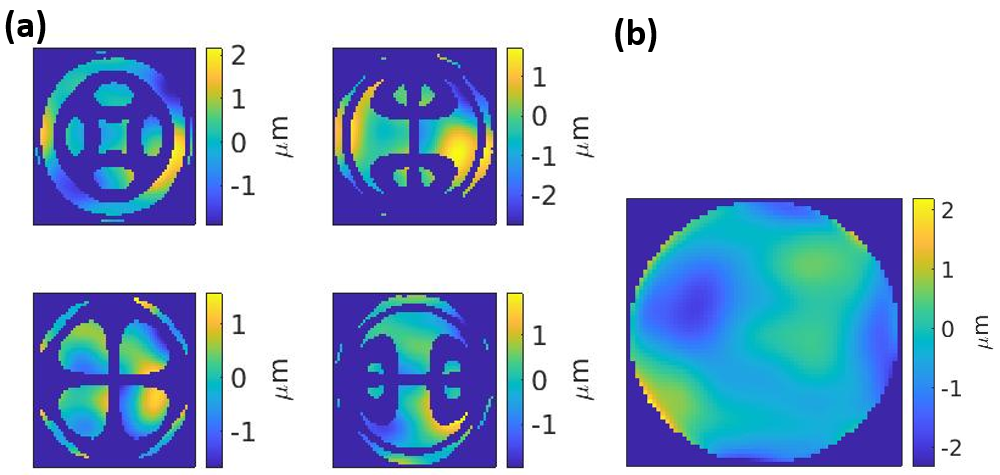}
    \caption{The difference of optical path length differences $\Delta L$ is shown in (a) and the corresponding difference topography $\Delta T$ is shown in (b). The aperture of the specimen has a radius of about $15$ mm.}\label{data sample}
\end{figure}

Each sample in the generated data set consists of a set of differences of optical path length differences $\Delta L$ and the difference topography $\Delta T$ between the specimen and the design topography. An example is shown in Figure \ref{data sample}. The difference of optical path length differences consists of four images, because four disjoint masks are sequentially used on the 2D point source array. In total, almost $40,000$ (virtual) topogaphies are generated for training and about $2,000$ are generated for testing. The mean root mean squared deviation to the design topography of the generated topographies in the test data set is $564$ nm and its median deviation is $473$ nm. The topographies of the test set range from a peak to valley difference of $42$ nm to $11.6\ \mu$m. Some more examples are shown in the Appendix \ref{appendix results} (Figure \ref{topographies}) to illustrate the diversity of the test data.

In \cite{Hoffmann2020}, the training data were generated without including reference planes to the 
model of the optical system, and simulated data were considered constructed under the assumption of a perfect model for the optical system. In this paper, systematic investigations on the impact of calibration errors are carried out. For this purpose, the test data are generated using a non-perfect optical system by adding deliberately calibration errors to the optical system. These errors represent errors caused by an imperfect calibration and will be termed \textit{calibration errors} in the following. To construct test data containing such calibration errors, the virtual reference planes of the perfectly calibrated optical system are systematically modified, which alters the beam path through the optical system, cf. Figure \ref{reference planes}. Recall that the training data were determined through simulating data for the perfectly calibrated optical system. 

Investigating the prediction of the trained net for test data corrupted by calibration errors allows the generalizability of the trained net to be explored for a more realistic scenario. From the point of view of machine learning the calibration errors used for constructing the test set imply testing the trained net on an out-of-distribution test set. We are particularly interested in the behavior of calculated uncertainties and to the extent to which they reflect the errors in the reconstructed topographies caused by the calibration errors. In total, the introduced calibration error affects the differences of optical path length differences $\Delta L$ up to a root mean squared deviation of $219$ nm on the test data set. We refer to Appendix \ref{appendix calibration} for further details about the construction of the test data.

Noisy test data are generated by adding Gaussian noise to the input of the existing test data. Note that the training set is fixed and does not adapt to errors introduced on the test set.

\section{Method}\label{section method}
The inverse problem at hand can be stated as follows. Find a map $f$, such that the difference of optical path length differences $\Delta L$ maps to the corresponding difference topography $\Delta T$ (see Fig. \ref{data sample}), i.e.:\begin{equation}
    f:\mathbb{R}^{4\times D\times D} \rightarrow \mathbb{R}^{D\times D},\ \Delta L \mapsto \Delta T,
\end{equation} where $D$ is the given or chosen resolution of the images. Here, we choose $D$ equal to $64$. 
The choice in the dimensionality of the optical path length differences and difference topographies is not mandatory and done for convenience here. The function $f$ can be approximated by a parameterization $f_\phi$ with parameter space $\Phi$ solving the following minimization problem for all possible tuples $(\Delta L,\Delta T)$:\begin{equation}
    \underset{\phi\in\Phi}{min}\Vert f_\phi (\Delta L) -\Delta T\Vert_2.
\end{equation}

Recall that the difference topography $\Delta T$ is the difference between the known design topography and the specimen at hand. Equally, the difference of optical path length differences $\Delta L$ is the difference of the optical path length difference (i.e. difference between optical path length of the ``measurement arm''and the ``reference arm'') derived from the design topography and the one measured based on the specimen. 
In a real TWI measurement the optical path length differences of the specimen are measured (based on the observed CCD values), while the optical path length differences of the design topography are calculated using the computer model of the optical system. For reasons of simplification $\Delta T$ and $\Delta L$ will from now on be simply called topography and optical path length difference, respectively, as there is a unique dependency.

\begin{figure}[t]
    \centering
    \includegraphics[scale = 0.45]{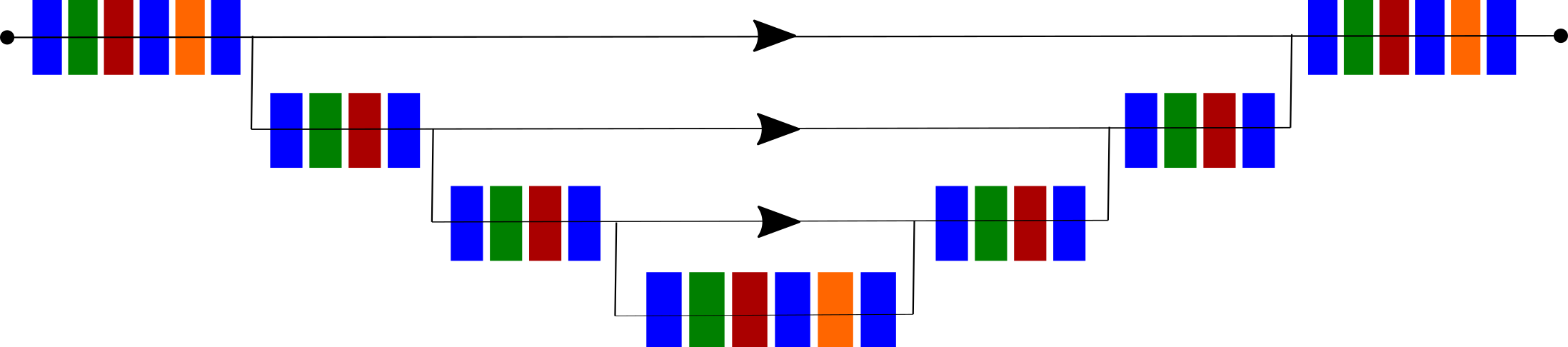}
    \caption{Example U-Net architecture.}\label{unet}
\end{figure}

Neural Networks have become a popular method used in imaging after the introduction of convolutional neural networks \cite{LeCun1995}. They can extract location invariant features and share weights which means that less parameters need to be trained. Deep neural networks, which are neural networks with many hidden layers, are commonly used today thanks to convolutional layers and technological advancement. The U-Net \cite{Ronneberger2015} is a specific deep neural network architecture which achieved great results in various imaging tasks (cf. \cite{Barbastathis2019}, \cite{Esser2018} or \cite{Icsil2019}). Therefore, we chose the U-Net as network architecture similar to \cite{Hoffmann2020}. An example of the network structure is shown in Figure \ref{unet}. The given input is processed through several mainly convolution layers and results in the predicted output from left to right. The network has a bottleneck structure which means that the dimension of the image data is reduced after each bundle of layers until it attains its minimal dimension in the center and increases afterwards in the same way. Furthermore, the arrows indicate skip connections between the bundles of layers of the same dimension which means that the output of the last layer with the same dimension is depth concatenated to the input after dimensional increase. 

It is well-known that trained neural networks typically find only a local and not a global optimum. Training multiple networks and making the prediction a decision of the constructed network ensemble \cite{Hansen1990} is a straightforward solution to overcome this problem to a certain extent. Other advantages of network ensembles are the simplicity of implementation and the scalability to high dimensional data if a well suited architecture can be found. We use an ensemble of U-Nets.

Furthermore, deep ensembles have been proposed in \cite{Lakshminarayanan2017} for uncertainty quantification because of their predictive variety. The random initialization, along with the random shuffling of data points while training, is considered to induce sufficient diversity to the network ensemble. We propose to also include dropout layers \cite{Baldi2013} during training to encourage diversity. In contrast to \cite{Lakshminarayanan2017}, we focus on model uncertainty and do not predict an extra variance per output neuron which would lead to many more learnables because of the high dimensional output.

Eight U-Nets are independently trained for the ensemble using the mean squared error as loss function and a $L2$ regularization with regularization factor $0.002$. Each network consists of $69$ layers including five max pooling layers and transposed convolution layers, respectively. The initial learning rate equals $5e-5$ with a learning rate drop factor of $0.75$ every fourth epoch. In total, each network trains for $25$ epochs with a mini batch size of $64$ and the data are randomly shuffled every epoch.

To sum up, we train an ensemble of $M=8$ deep neural networks $\lbrace f_{\phi_1},\ldots,f_{\phi_M}\rbrace$, $\phi_1,\ldots,\phi_M\in\Phi$, that all have the same U-Net architecture and are independently trained on the same training data. Diversity is induced through random initialization, random data shuffle per epoch and dropout layers included during the training procedure. The ensemble prediction is defined as the average over all predictions, i.e.: \begin{equation}
    f_\phi(\Delta L) := \frac{1}{M}\sum_{j=1}^Mf_{\phi_j}(\Delta L)\ \in\mathbb{R}^{D\times D}.
\end{equation}
We define the \textit{ensemble uncertainty} as the standard deviation over the predictions: \begin{equation}\label{eq quantile}
    uc(f_\phi(\Delta L)) := \left(\frac{1}{M}\sum_{j=1}^M\left(f_{\phi_j}(\Delta L)-f_\phi(\Delta L)\right)^{.2}\right)^{.\frac{1}{2}} \in\mathbb{R}^{D\times D} \,.
\end{equation}
In \eqref{eq quantile}, $^{.2}$ and $^{.}\frac{1}{2}$ indicate elementwise square and square root, respectively. This definition is in line with the uncertainty definition from \cite{Lakshminarayanan2017} when omitting the aleatoric part. If the ensemble uncertainty is considered on an entire topography, we refer to the \textit{topography uncertainty} defined as $\sqrt{ \frac{1}{D'}\sum_{d=1}^{D'} uc(f_\phi(\Delta L))_d^2} \in\mathbb{R}$, where $D'$ is the number of pixel coordinates of the predicted topography $\Delta T$.

\section{Results} \label{section results}
This section presents the results obtained when considering a test set constructed using a perfectly calibrated system and test data produced by an optical system containing calibration errors. In addition, results are shown when the input data from the test set are corrupted by noise. Recall that the training data are constructed by simulating a perfectly calibrated optical system. Particularly the test set constructed after introducing calibration errors into the physical model used to generate the data can be seen as an out-of-distribution test and explores the generalizability of the trained network. Our focus lies on analyzing the uncertainty quantification produced by the network ensemble. As mentioned above, the uncertainty mainly estimates the epistemic uncertainty and not the aleatoric uncertainty \cite{Gal2017}, as there is no noise in the simulated training data and because of the high dimensional output. Nonetheless, the behavior of the ensemble uncertainty is analyzed on out-of-distribution test data through the systematically introduced calibration errors, and also through analyzing test data whose input is corrupted by white noise.

\subsection{Perfectly calibrated system}
\begin{figure}[t]
  \centering
  \begin{minipage}[t]{0.56\textwidth}
    \centering
    \includegraphics[scale = 0.65]{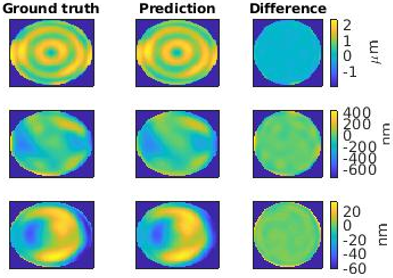}
    \caption{The ground truth, the ensemble prediction and their difference are shown for three example topographies with a median error of $44.5$ nm, $17.2$ nm and $1.5$ nm, respectively.}\label{predictions}
  \end{minipage}
  \hfill
  \begin{minipage}[t]{0.41\textwidth}
    \centering
    \includegraphics[scale = 0.75]{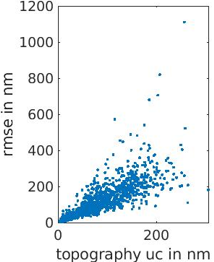}
    \caption{The root mean squared errors of the ensemble predictions of the topographies are plotted against their corresponding topography uncertainties.}\label{uc vs rmse}
  \end{minipage}
\end{figure}

The performance of the trained network ensemble is evaluated on the test data set generated by the same optical system as the training data, i.e. a perfectly calibrated system. The average root mean squared error equals $77$ nm. First results of the ensemble prediction are shown in Figure \ref{predictions}. Three example topographies with greatly varying size are reconstructed and the difference between the ground truth and the prediction is given. The network ensemble has no problem reconstructing the different topographies. The main error occurs at the edge of the topographies. This is not surprising since the input data is here more error-prone at the topography edges. There exists almost no redundant information since the patches rarely overlap at the edges, and some rays even leave the optical system without returning to the CCD. It is well-known that outlier have a high impact on the $L2$ measure. Therefore, the median error (i.e. the median of the absolute errors) of an image is a more stable measure than the root mean squared error to capture the total predictive capacity of the network ensemble. Its average over the entire test set equals $20$ nm.

\begin{figure}[t]
  \centering
  \begin{minipage}[t]{0.58\textwidth}
    \centering
    \includegraphics[scale = 0.5]{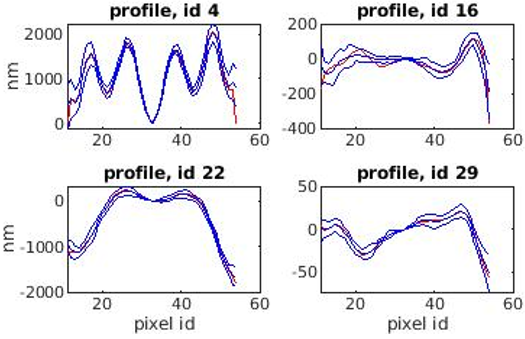}
    \caption{The profiles of four example topographies (cf. Figure \ref{topographies}) are plotted in red, along with the profiles of the ensemble predictions and the estimated uncertainty tubes in blue.}\label{profile}
  \end{minipage}
  \hfill
  \begin{minipage}[t]{0.39\textwidth}
    \centering
    \includegraphics[scale = 0.5]{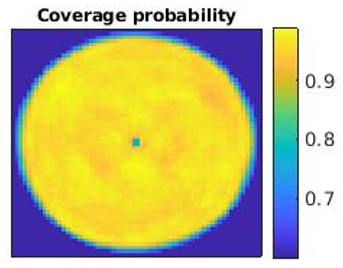}
    \caption{The uncertainty estimation does cover the prediction error well if the coverage probability is close to one.}\label{coverage probability}
  \end{minipage}
\end{figure}

The relationship between topography uncertainty and root mean squared error of the topography predictions by the ensemble is shown for the test data set in Figure \ref{uc vs rmse}. The uncertainty grows for increasing root mean squared error. A more detailed analysis of the uncertainty estimation is given in Figure \ref{profile}, where the profile of a topography is plotted along with the profile of the ensemble prediction and the estimated \textit{uncertainty tube}, i.e. the profiles of $\Delta T$, $f_\phi (\Delta L)$ and $f_\phi (\Delta L) \pm 1.96uc(f_\phi(\Delta L))$. The ground truth (in red) rarely leaves the uncertainty tube (in blue) and at the same time, the uncertainty tube is not too wide. The uncertainty tube is in general widest at the borders of the topography and smallest at its center. This behavior is in accordance with the corresponding sizes of rms errors. Some further examples are given in Appendix \ref{appendix results}, Figure \ref{profiles}.

The factor $1.96$ for the uncertainty tube equals the $97.5\%$ quantile of the standard normal distribution. This choice implies that if the errors are normally distributed around the prediction of the ensemble with a standard deviation equal to $uc(f_\phi(\Delta L))$, then the uncertainty tube will encompass the difference of predictions and ground truth in $95\%$ of the cases.

Figure \ref{coverage probability} shows the coverage probability of the uncertainty estimation. For each pixel $p$ of the topography the \textit{coverage probability} $cp$ is estimated as follows: \begin{equation}\label{eq cp}
    cp(p) = \frac{1}{N}\sum_{i=1}^Ng(u_i(p), y_i(p)), \ 
    g(u_i, y_i) :=
    \begin{cases}
        1, & if \frac{|y_i-\hat{y}_i|}{1.96u_i}\leq 1 \\
        0, & otherwise
    \end{cases}
    ,
\end{equation}
where $y_i$ is the ground truth topography height at pixel $p$ of the $i$th data sample, $\hat{y}_i$ is the corresponding predicted topography height and $u_i$ the uncertainty estimate as defined in \eqref{eq quantile}. The coverage probability indicates how likely it is that the ground truth is found around the prediction within the chosen uncertainty tube. Figure \ref{coverage probability} shows the pixelwise coverage probabilities indicating that the calculated uncertainties characterize well the size of the errors of predictions. The \textit{total coverage probability} can be defined as follows: \begin{equation}\label{eq total cp}
    \frac{1}{D'}\sum_{d=1}^{D'} \frac{1}{N}\sum_{i=1}^Ng(u_i(p_d), y_i(p_d))\ \in\mathbb{R}.
\end{equation} The observed total coverage probability equals $94\%$, which fits well the intended $95\%$ coverage probability.

Altogether, the network ensemble makes good predictions and provides a trustworthy uncertainty estimate, not only per image, but also pixelwise, for the perfectly calibrated optical system.

\subsection{Systematically introduced calibration error}
\begin{table}[h]
\centering
\begin{tabular}{|c|c@{\hspace{1.25\tabcolsep}}c@{\hspace{1.25\tabcolsep}}c@{\hspace{1.25\tabcolsep}}c@{\hspace{1.25\tabcolsep}}c@{\hspace{1.25\tabcolsep}}c@{\hspace{1.25\tabcolsep}}c@{\hspace{1.25\tabcolsep}}c@{\hspace{1.25\tabcolsep}}c@{\hspace{1.25\tabcolsep}}c@{\hspace{1.25\tabcolsep}}c|}
\hline
\begin{tabular}[c]{@{}c@{}}\textbf{calib error}\\ (in $\%$)\end{tabular} & $0$ & $10$ & $20$ & $30$ & $40$ & $50$ & $60$ & $70$ & $80$ & $90$ & $100$ \\ \hline
\begin{tabular}[c]{@{}c@{}}\textbf{flawed input}\\ (in nm)\end{tabular} & $0$ & $28$ & $55$ & $75$ & $96$ & $116$ & $137$ & $156$ & $176$ & $198$ & $219$ \\ \hline
\begin{tabular}[c]{@{}c@{}}\textbf{rmse single net}\\ (in nm)\end{tabular} & $95$ & $137$ & $178$ & $229$ & $280$ & $331$ & $383$ & $435$ & $484$ & $538$ & $596$ \\ \hline
\begin{tabular}[c]{@{}c@{}}\textbf{rmse ensemble}\\ (in nm)\end{tabular} & $77$ & $104$ & $145$ & $194$ & $241$ & $288$ & $336$ & $384$ & $429$ & $479$ & $536$ \\ \hline
\begin{tabular}[c]{@{}c@{}}\textbf{median error}\\(in nm)\end{tabular} & $20$ & $29$ & $42$ & $55$ & $68$ & $82$ & $95$ & $110$ & $125$ & $139$ & $154$ \\ \hline
\begin{tabular}[c]{@{}c@{}}\textbf{topography uc}\\ (in nm)\end{tabular} & $55$ & $90$ & $108$ & $131$ & $155$ & $178$ & $201$ & $225$ & $246$ & $267$ & $285$ \\ \hline
\begin{tabular}[c]{@{}c@{}}\textbf{total $cp$}\\ (in $\%$)\end{tabular} & $94$ & $94$ & $84$ & $78$ & $75$ & $73$ & $71$ & $70$ & $69$ & $68$ & $67$ \\ \hline
\end{tabular}\caption{The influence of the introduced calibration error is analyzed on the test data set. The rows present the following values: the percentage of induced calibration error, its impact on the input data expressed as the root mean squared error, the averaged root mean squared error over the single network predictions, the root mean squared error of the ensemble prediction, the median absolute error of the ensemble prediction, the topography uncertainty and the total coverage probability.}\label{table results}
\end{table}

In the last subsection training and test set were generated through the same optical system. However, the network ensemble should also make trustworthy predictions on out-of-distribution data. Indeed, in any real measurement scenario there will remain a calibration error. Therefore, the quality of the ensemble prediction and its uncertainty quantification are analyzed under the influence of a systematically introduced, growing calibration error. To this end, the optical system generating the test data is increasingly deviated from the optical system used for producing the training data. The chosen topographies in the test set remain the same (Section \ref{section data generation}).

\begin{figure}[h!]
  \centering
  \begin{minipage}[t]{0.57\textwidth}
    \centering
    \includegraphics[scale = 0.66]{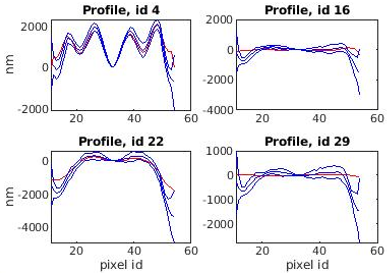}
    \caption{The profiles of four example topographies (Fig. \ref{topographies}) are plotted (red), along with the ensemble predictions and the estimated uncertainty tubes (blue) under full calibration error.}\label{profile_c1}
  \end{minipage}
  \hfill
  \begin{minipage}[t]{0.41\textwidth}
    \centering
    \includegraphics[scale = 0.7]{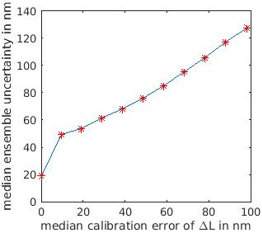}
    \caption{The median of the ensemble uncertainty on the test data set is plotted against the median of the absolute calibration error on the input data $\Delta L$.}\label{cerror_vs_uc}
  \end{minipage}
\end{figure}

A first overview of results is summarized in Table \ref{table results}. The calibration error is induced stepwise from zero to a hundred per cent, cf. first row. The second row shows the actual impact of the corrupted optical system on the resulting optical path length differences which are taken as input data by the trained net for predicting the sought topography. Then, the root mean squared error of the ensemble prediction is given in the fourth row which achieves consistently better results than the prediction of a single trained network (third row). A more robust measure for the ensemble prediction is the median error in the fourth row, because it is more stable against outliers at the edges of the topographies. The fifth row displays the mean topography uncertainties. Finally, the total coverage probabilities are calculated in the last row (cf. eq. \eqref{eq total cp}).

Analogous to Figure \ref{profile}, the profiles of the four test topographies are plotted (in red) in Figure \ref{profile_c1}, along with their ensemble predictions and their estimated uncertainty tubes (in blue) for the full calibration error. In the first column, the ground truth topographies mostly are within the range defined by the predicted topographies and their calculated uncertainties. Furthermore, the ensemble predictions recognize the basic shapes of the sought topographies, except for the edges. In contrast, the topographies from the second column are not recognized well. Instead, the ensemble predictions resemble one another and are both predicting a topography with much stronger peak to valley variability. This prediction behavior can be explained when considering the ground truth. The impact of the calibration error dominates the ensemble prediction for small topographies, while it has a smaller impact on larger difference topographies. More examples and the step wise change in the ensemble prediction for increasing calibration error are shown in the Appendix Figures \ref{profiles_c1} and \ref{profilesall_c}, respectively.

\begin{figure}[t]
 \centering
 \includegraphics[scale = 0.8]{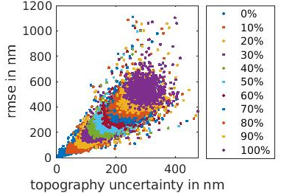}
 \caption{The root mean squared error of the ensemble prediction is plotted against the topography uncertainty. Each colour represents a different amount of introduced calibration error.}\label{uc_vs_rmse_c_all}
\end{figure}

The dependency between the ensemble uncertainty and the increasing calibration error is plotted in Figure \ref{cerror_vs_uc}. The ensemble uncertainty grows with growing calibration error, which is a desired behavior for trustworthy predictions. Figure \ref{cerror_vs_uc} shows the results over the entire test data set while Figure \ref{profile_c1} presents the pixelwise results. In contrast, Figure \ref{uc_vs_rmse_c_all} shows the obtained results on the image level, where the root mean squared error of the ensemble prediction is plotted against the topography uncertainty for the different degrees of induced calibration error. Again, the estimated uncertainty correlates well with the prediction error (as well as with the calibration error).

\begin{figure}[t!]
    \centering
    \includegraphics[scale = 0.53]{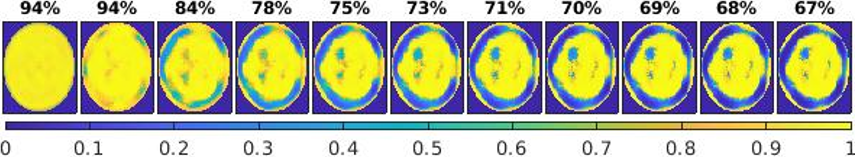}
    \caption{The coverage probabilities per pixel are plotted together with their total coverage probability on the test set for increasing calibration error (from left to right).}\label{coverage_probabilities}
\end{figure}

Finally, the coverage and total coverage probabilities (eq. \ref{eq cp}-\ref{eq total cp}) are shown in Figure \ref{coverage_probabilities} in relation to the growing calibration error. The coverage probability slowly decreases. Nonetheless, the total coverage probability stays at $94\%$ after having induced $10\%$ of the calibration error and still correctly covers two thirds of the pixels for the maximal induced corruption of the input data in the test set. Furthermore, not the center but the topgoraphy edges are less well covered with increasing calibration error.

To sum up, ensemble prediction and its uncertainty quantification are best when the optical system is perfectly calibrated and get worse with an increasingly worse calibration. However, the uncertainty increases with growing calibration error and appears to still characterize reliably the size of the errors in the predictions.

\subsection{Noisy data}
The previous subsection examines the influence of systematically deviating the computer model of the optical system, that is used to simulate the test data, on the network ensemble. Another source of error is noise in the data. Therefore, the ensemble prediction and its uncertainty estimation are analyzed in the following using noisy input data. Again, only the test data are modified while the trained ensemble stays fixed.

\begin{figure}[t]
    \begin{minipage}[t]{0.46\textwidth}
    \centering
    \includegraphics[scale = 0.9]{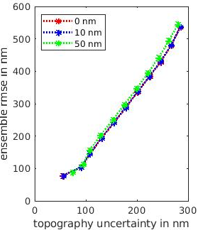}
    \caption{The averaged root mean squared error of the ensemble predictions is plotted against the averaged topography uncertainties under the influence of varying Gaussian noise (legend shows standard deviations) and an increasing calibration error.}\label{compare_results_noise}
  \end{minipage}
  \hfill
  \centering
  \begin{minipage}[t]{0.52\textwidth}
    \centering
    \includegraphics[scale = 0.83]{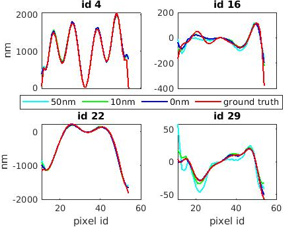}
    \caption{The ensemble predictions and their ground truth are plotted for four different test topographies under the influence of varying Gaussian noise (legend shows standard deviations) for the perfectly calibrated system.}\label{noise_ideal_case}
  \end{minipage}
\end{figure}

Random white noise is added pixelwise to the input data with a standard deviation of $10$ nm and $50$ nm, respectively. This is done for the test data generated by the perfectly calibrated optical system as well as for the test data produced by the optical system after corrupting it with an increasing calibration error. Figure \ref{compare_results_noise} shows the results averaged over the entire test set. The root mean squared error of the ensemble predictions is plotted against its corresponding estimate of topography uncertainties for the stepwise increasing introduced calibration error and the input data additionally corrupted by normally distributed noise with a standard deviation of $10$ nm (blue), a standard deviation of $50$ nm (green) and without noise (red). There is no perceptible difference between the perturbed data of $10$ nm standard deviation and the clean data without noise. This is true for the perfectly calibrated case as well as for all systematically introduced calibration errors. The prediction errors are only slightly greater even for the Gaussian noise with a standard deviation of $50$ nm.

\begin{figure}[t]
    \centering
    \includegraphics[scale = 0.7]{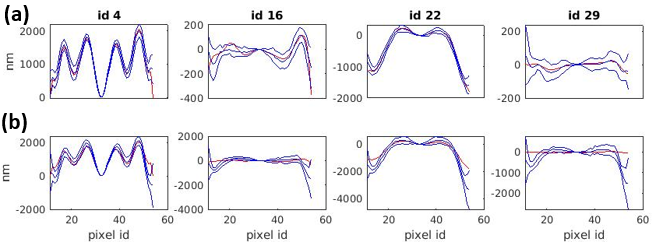}
    \caption{The profiles of the ensemble prediction and uncertainty tube are plotted (blue) with the underlying ground truth (red) for perturbed input data by white noise with a standard deviation of $50$ nm for (a) the perfectly calibrated system and (b) the full calibration error.}\label{profiles_noise}
\end{figure}

A closer look at some example topographies is given in Figures \ref{noise_ideal_case} and \ref{profiles_noise} where the profiles are plotted. The former shows the ground truth together with the ensemble prediction for the perfectly calibrated data with and without noise. The noisy data has almost no impact on the ensemble prediction for the large topographies in the first row. In contrast, especially the noisy data with a standard deviation of $50$ nm have a visible influence on the ensemble prediction of the smaller topography at the bottom right. Nonetheless, the main shape of the topography is still recognized.

Figure \ref{profiles_noise} shows the ground truth along with the ensemble prediction and estimated uncertainty for the noisy input data with standard deviation of $50$ nm. The first row (a) shows the results for the perfectly calibrated optical system and can be compared to Figure \ref{profile}. Here, the uncertainties are much larger for the small topography (id $29$) under the influence of noise which is an appropriate behavior as the prediction is also most influenced by the noise for small topographies (Fig. \ref{noise_ideal_case}). The second row (b) shows the same plots but for the maximal calibration error which seems to dominate the noise and there is no visible difference to Figure \ref{profile_c1}.

In total, the network ensemble is not largely affected by noisy data although it has not seen them while training. The prediction error might grow in some cases, but then the uncertainty estimate rises as well.

\section{Discussion and conclusion}\label{section conclusion}
In this paper we show that ensemble learning is a powerful tool to reliably solve the inverse problem of reconstructing the topographies from given optical path length differences. Moreover, the proposed ensemble method provides a sensible uncertainty quantification to its results, which is shown on the pixel level as well as the image level. This is true not only for a diverse test set that is disjoint from the training data, but also under the influence of different additional error sources. The network ensemble is successfully tested on out-of-distribution data realized by systematically deviating the computer modeled optical system used for simulating the test data, which corresponds to an increasing calibration error. The estimated uncertainty grows in relation with the introduced calibration error and the prediction error, respectively. Furthermore, the ensemble behavior is analyzed under the influence of noisy input data using different amounts of white noise. The noisy input does effect the prediction almost not at all. An impact can be seen mainly for small topographies. However, the uncertainty estimation is able to reflect this behavior as well.

From an application point of view, ensemble learning can be used to reliably solve the inverse topography reconstruction problem up to a certain precision much faster than applied conventional methods after having trained the neural networks once for a specific design topography. Additionally, the network ensemble provides a consistent uncertainty quantification. Including other error sources such as positioning errors of the specimen, or applying the proposed method to real data could be the next steps. Ensemble learning could also be applied to validate the current status of a calibration using a fixed reference specimen. Even if the topography is not perfectly known, the ensemble uncertainty should increase if the calibration worsens over time. 

From a machine learning perspective, this application exemplifies the ability of ensemble methods to make trustworthy predictions and to provide an uncertainty quantification. The great potential of ensemble methods lies in their simple and straightforward implementation when solving high dimensional problems. The proposed uncertainty quantification mainly considers epistemic uncertainty; future work could address the explicit incorporation of aleatoric uncertainty sources as well. Also, establishing a high dimensional benchmark data set to test and compare scalable uncertainty methods is referred to as future work.

\section*{Acknowledgement}
The authors thank Manuel Stavridis for providing the simulation toolbox SimOptDevice and Michael Schulz for fruitful discussions about optical form measurements.
\bibliography{ref}{}
\bibliographystyle{abbrv}

\appendix
\section{Asphere}\label{appendix asphere}
The asphere used as design topography for the data generation process is quantified as follows. The aspherical coordinates $(A_4,A_6,\ldots,A_{16})^T$ are $(5.4145e+03\ m^{-3}, -8.0413e+05\ m^{-5}, -2.9871e+09\ m^{-7}, -1.4918e+12\ m^{-9}, 1.3777e+15\ m^{-11}, 4.4258e+18\ m^{-13}, -3.4928e+21\ m^{-15})^T$, the conic constant $\kappa$ equals $-1$ and the paraxial surface radius $R$ equals $0.0202$ m. The asphereical equation is given in \cite{Braunecker2008} (2.2.2.1).

\section{Reference planes}\label{appendix calibration}
The light path through the optical system is deviated by introducing the virtual reference planes $R_1$ and $R_2$ before and after the topography, respectively (Fig. \ref{reference planes}). Note, that two reference planes are required to ensure that the calibration is valid regardless the surface under test.

Each reference plane is parameterized by a double fit of Zernike polynomials \cite{Baer2014_2}. The parameterization of the \textit{source} reference plane $R_1$ is depending on the intersection $(u,v)$ with a beam and its originating light source $(U,V)$, i.e.: \begin{equation}
    L_{R_1}(u,v,U,V) = \sum_{i=1}^I\left(\sum_{j=1}^JQ_{ij}Z_j(U,V)\right)Z_i(u,v) ,
\end{equation} where $(Q_{ij})_{ij}$ is the matrix of Zernike coefficients with dimension $I\times J$. Analogously, the \textit{pixel} reference plane $R_2$ depends on the intersection $(m,n)$ with a beam and its corresponding pixel on the CCD $(M,N)$, i.e.:\begin{equation}
    L_{R_2}(m,n,M,N) = \sum_{k=1}^K\left(\sum_{h=1}^HP_{kh}Z_h(M,N)\right)Z_k(m,n) ,
\end{equation} where $(P_{kh})_{kh}$ is the matrix of Zernike coefficients with dimension $K\times H$. The total optical path length difference can then be computed by adding the optical path length to the source reference plane $L_{R_1}$ with the optical path length from the pixel reference plane to the CCD $L_{R_2}$ and the optical path length covered between the two reference planes $L_T$ depending on the topography $T$ and finally, substracting the optical path length coming from he reference arm $L_R$ (Fig. \ref{twi}: \begin{align}
    \nonumber L(u,v,m,n,U,V,M,N,T) = & L_{R_1}(u,v,U,V) + L_{R_2}(m,n,M,N) \\ 
    &+ L_T(u,v,m,n,T) - L_R(M,N).
\end{align}

The Zernike coefficients of the parameterized reference planes are iteratively adjusted during the calibration procedure \cite{Baer2014_2}. This is done in such a way that the measurements through the computer modeled optical system resemble more and more the measurements obtained from the real optical system for some well-known spherical calibration specimens. The calibration error is chosen to demonstrate effects of out-of-distribution data and does not necessarily reflect real world calibration errors \cite{Schindler2020}.

\section{Additional plots}\label{appendix results}
\begin{figure}[h]
    \centering
    \includegraphics[scale = 0.9]{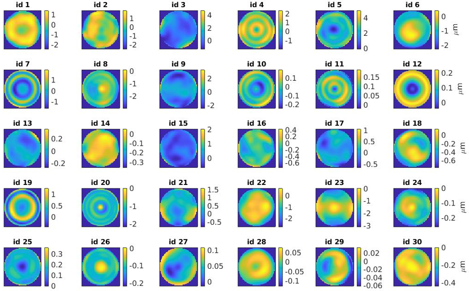}
    \caption{Some examples of generated test topography deviations $\Delta T$ are shown. The aperture has a radius of about $15$ mm.}\label{topographies}
\end{figure}

\begin{figure}[h]
    \centering
    \includegraphics[scale = 0.675]{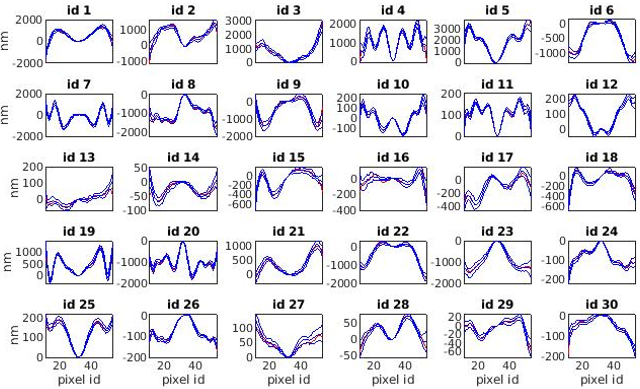}
    \caption{The profile plots of all topographies from Figure \ref{topographies} are plotted in red, while the ensemble predictions and the estimated uncertainties are plotted in blue.} \label{profiles}
\end{figure}

\begin{figure}[h!]
    \centering
    \includegraphics[scale = 0.69]{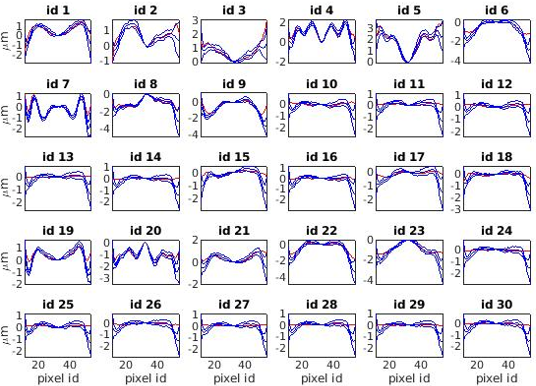}
    \caption{The profile plots of all topographies from Figure \ref{topographies} are plotted in red, while the ensemble predictions and the estimated uncertainties are plotted in blue for the full calibration error.} \label{profiles_c1}
\end{figure}

\begin{figure}[t]
    \centering
    \includegraphics[scale = 0.875]{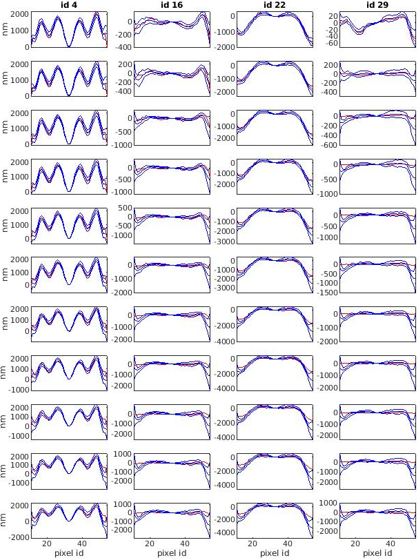}
    \caption{The profiles are plotted of the ensemble prediction and its estimated uncertainty tube (blue) together with the ground truth (red) for four difference topographies. The first row shows the results for the perfectly calibrated system. Then, the calibration error is increased by $10\%$ in each row.} \label{profilesall_c}
\end{figure}

\begin{figure}[h]
    \centering
    \includegraphics[scale = 0.662]{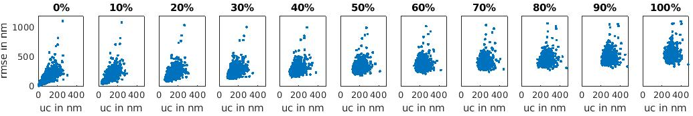}
    \caption{The root mean squared errors are plotted against the topography uncertainties of all data points in the test set under the influence of the systematically growing calibration error from left to right.}\label{uc_vs_rmse_c}
\end{figure}

\end{document}